\documentclass{article}

\usepackage{style}
\usepackage{multirow}
\usepackage{array}

\usepackage[utf8]{inputenc} 
\usepackage[T1]{fontenc}    
\usepackage{hyperref}       
\usepackage{url}            
\usepackage{booktabs}       
\usepackage{amsfonts}       
\usepackage{nicefrac}       
\usepackage{microtype}      
\usepackage{lipsum}
\usepackage{fancyhdr}       
\usepackage{graphicx}       
\graphicspath{{media/}}     
\usepackage{algorithm}
\usepackage{algpseudocode}
\usepackage{amsmath}

\usepackage{graphicx}
\title{Advancing Arabic Speech Recognition Through Large-Scale Weakly Supervised Learning}

\author{
\begin{tabular}{c}
    Mahmoud Salhab, Marwan Elghitany, Shameed Sait, \\
    Syed Sibghat Ullah, Mohammad Abusheikh, and Hasan Abusheikh
    \end{tabular} \\
  CNTXT AI \\
  AI Department \\
  Abu Dhabi, UAE\\
  \texttt{\{{mahmoud.salhab,marwan.elghitany,shameed.ali,syed.sibghat,mas,has}\}@cntxt.tech} \\
}

\begin{document}
\maketitle

\begin{abstract}
Automatic speech recognition (ASR) is crucial for human-machine interaction in diverse applications like conversational agents, industrial robotics, call center automation, and automated subtitling. However, developing high-performance ASR models remains challenging, particularly for low-resource languages like Arabic, due to the scarcity of large, labeled speech datasets, which are costly and labor-intensive to produce. In this work, we employ weakly supervised learning to train an Arabic ASR model using the Conformer architecture. Our model is trained from scratch on 15,000 hours of weakly annotated speech data covering both Modern Standard Arabic (MSA) and Dialectal Arabic (DA), eliminating the need for costly manual transcriptions. Despite the absence of human-verified labels, our approach achieves state-of-the-art (SOTA) results in Arabic ASR, surpassing both open and closed-source models on standard benchmarks. By demonstrating the effectiveness of weak supervision as a scalable, cost-efficient alternative to traditional supervised approaches, paving the way for improved ASR systems in low resource settings.
\end{abstract}


\section{Introduction}

Automatic speech recognition (ASR), also known as speech-to-text (STT), refers to the process of converting an input speech signal into readable text \cite{10.1007/978-3-030-21902-4_2}. It is a fundamental component of human-machine interaction, enabling users to communicate with machines through spoken language, thereby allowing systems to act based on verbal commands \cite{kheddar2024automatic}. As a result, ASR systems are now widely deployed across various domains, including but not limited to healthcare and medical voice assistance, industrial robotics, law enforcement, criminal justice, crime analysis, telecommunications, smart home automation, security access control, and consumer electronics \cite{vajpai2016industrial, huang1991study}. Among the languages driving this technological evolution, Arabic stands out as a critical yet often under-researched area. As the fourth most used language on the internet and one of the six official languages of the United Nations, Arabic serves as the primary communication medium for millions of people across 22 countries \cite{6841973}. Therefore, the development and refinement of robust ASR systems tailored to the complexities of the Arabic language are of paramount importance.

Linguistically, Arabic can be classified into three distinct forms: (i) Classical Arabic (CA), the language of historical and religious texts, most notably the Holy Quran, which remains widely understood despite its archaic structure; (ii) Modern Standard Arabic (MSA), a contemporary adaptation of CA that is standardized and used in formal communication, media, and education; and (iii) Dialectal Arabic (DA), which comprises regionally diverse spoken varieties that differ significantly from both CA and MSA \cite{ALAYYOUB2018522}.

Despite the widespread use of Arabic, it remains a low-resource language in the field of ASR. The scarcity of publicly available transcribed datasets has hindered advancements in Arabic ASR. In response, several efforts have been made to bridge this gap by curating and releasing Arabic speech datasets, such as MASC \cite{e1qb-jv46-21} and SADA \cite{10446243}, among others. While these datasets have significantly contributed to the progress of Arabic ASR and facilitated experimentation, they remain limited in both size and linguistic diversity, restricting the generalization of Arabic ASR models. On the other hand, neural network-based ASR systems typically require extensive amounts of transcribed speech data to achieve high performance \cite{lu2020exploring, 9414087}. However, large-scale manual transcription and dataset curation are both time-consuming and labor-intensive, making data acquisition a major challenge \cite{gao2023learning}. To overcome this challenge, we propose a weakly supervised learning approach for training an Arabic ASR model based on the Conformer architecture \cite{gulati2020conformer}. Our model is trained on a large corpus of weakly annotated speech data covering both Modern Standard Arabic (MSA) and Dialectal Arabic (DA), eliminating the need for costly manual transcriptions. Despite the absence of human-verified labels, our method achieves state-of-the-art (SOTA) performance, surpassing all previous Arabic ASR models on the standard benchmarks. This demonstrates the effectiveness of weak supervision in developing high-performance ASR systems for low-resource languages.

This paper is organized as follows: Section \ref{sec:related-work} presents a comprehensive literature review covering ASR, weakly supervised learning, and Arabic ASR research. Section \ref{sec:methodology} outlines our proposed methodology, followed by Section \ref{sec:experiments} detailing the experimental design, evaluation methodology, and presents the resulting empirical findings. Finally, Section \ref{sec:conclusion} summarizes our findings and discusses potential future research directions.

\section{Related Work}
\label{sec:related-work}

Automatic Speech Recognition (ASR) for Arabic remains a challenging task due to several difficulties, including data sparsity, lexical variety, morphological complexity, and the presence of multiple dialects across the Arab world \cite{ali2014advances, Cardinal2014, diehl2012morphological}. Traditional ASR systems typically follow a cascaded approach, integrating components such as Gaussian Mixture Models (GMMs), Hidden Markov Models (HMMs), and Deep Neural Networks (DNNs) to build hybrid HMM-DNN architectures.

Several studies have explored conventional ASR approaches for Arabic. For instance, \cite{Cardinal2014} developed an Arabic broadcast news ASR system using the Kaldi toolkit, trained on 50 hours of Al-Jazeera news content, demonstrating that the best-performing approach was the hybrid DNN-HMM model. Similarly, \cite{bouchakour2018improving} investigated enhancing Arabic ASR performance in mobile communication systems by leveraging HMM-DNN models. Another study \cite{hussein2022arabic} experimented with an encoder-decoder transformer model alongside a TDNN-LSTM language model for Arabic ASR.

Given the complexity of Arabic dialects, several studies have focused on dialectal ASR. Researchers in \cite{10.1109/ICASSP.2013.6639311} developed an ASR system specifically for Egyptian Arabic. Meanwhile, \cite{MENACER201781} introduced a DNN-HMM-based ASR system tailored for Algerian Arabic alongside Modern Standard Arabic (MSA) in addition to using n-gram language models. To overcome data limitations, \cite{khurana2019darts} proposed the DARTS ASR system for Egyptian Arabic, employing transfer learning by initially training the acoustic model on broadcast data and later fine-tuning it on Egyptian Arabic data.

Recent advancements in ASR have shifted towards end-to-end models. Researcher in \cite{graves2014towards} introduced an end-to-end model capable of directly mapping acoustic features to text. Additionally, \cite{https://doi.org/10.1049/sil2.12057} proposed a CTC-based ASR model using a CNN-LSTM alongside an attention-based approach to improve Arabic ASR, particularly for diacritized text.

Due to the scarcity of high-quality ASR training datasets, researchers have explored alternative training methodologies. One approach involves leveraging weakly supervised learning, as demonstrated in \cite{pmlr-v202-radford23a}, where ASR was trained on a large-scale weakly supervised dataset containing 680,000 hours of speech, resulting in a model that generalizes well across standard benchmarks on various datasets. Other strategies include self-supervised learning \cite{baevski2020wav2vec}, transfer learning \cite{kheddar2023deep}, and weakly supervised training \cite{gao2023learning}. Moreover, studies such as \cite{galvez2021people, chen2021gigaspeech} have proposed automated pipelines to expand ASR training datasets.

Finally, research suggests that ASR models trained on diverse datasets across multiple domains exhibit greater robustness compared to those trained on a single dataset. This has been highlighted in \cite{narayanan2018toward, likhomanenko2020rethinking}, where multi-domain training was shown to enhance generalization and performance across different ASR benchmarks.

\section{Methodology}
\label{sec:methodology}

\subsection{Weakly Supervised Learning}

\begin{figure}[htbp]
    \centering
    \includegraphics[width=\linewidth]{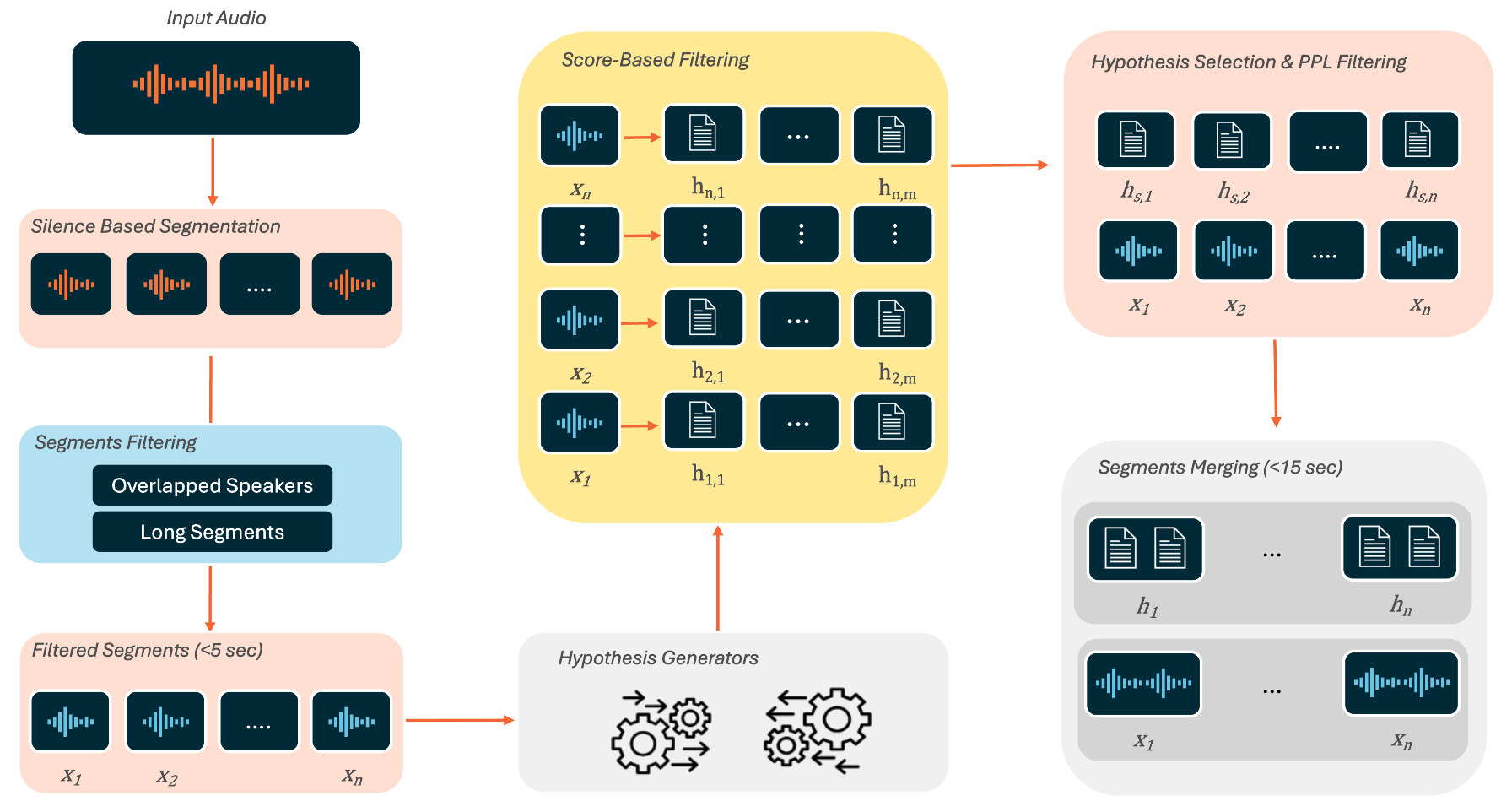}
    \caption{Illustration of the weak label generation process, which includes speech processing, generating multiple hypotheses, selecting the most probable hypotheses, and segments merging.}
    \label{fig:pipeline}
\end{figure}

Formally, in supervised training for automatic speech recognition (ASR) models, we define a training dataset $\mathcal{D}_{tr}$ and a test dataset $\mathcal{D}_{te}$ as follows:  
\[
\mathcal{D}_{tr} = \{(x_1, y_1), (x_2, y_2), \dots, (x_N, y_N)\}, \quad  
\mathcal{D}_{te} = \{(x_1, y_1), (x_2, y_2), \dots, (x_M, y_M)\}.
\]  
The objective is to learn a mapping function $\mathcal{F}_{\theta}: \mathcal{X} \mapsto \mathcal{Y}$, where $\mathcal{X}$ represents the input features. In the context of ASR, $\mathcal{X}$ corresponds to an audio representation, and in this work, it is specifically a mel-spectrogram of the input audio signal. Formally, $\mathcal{X} \in \mathbb{R}^{T \times F}$, where $T$ denotes the time dimension and $F$ represents the number of mel-frequency bins.

The output space $\mathcal{Y}$ consists of the target sequence, which may be represented as sequence of characters or subword tokens:  
\[
\mathcal{Y} = \{s_1, s_2, \dots, s_J\},
\]  
whereby, each token $y_{i,j}$ is an element of a vocabulary $\mathbb{V}$. Given training samples $x_i \in \mathcal{X}$ and their corresponding labels $ \quad y_i = \{y_{i,j} \mid j = 0, 1, \dots, J\}$, such that $\forall j, \, y_{i,j} \in \mathcal{Y}$, the supervised learning setup assumes that target label are correctly annotated and free from noise, typically provided by human. On top of that, each training instance $(x_i, y_i)$ is assumed to be independently and identically distributed (i.i.d.) according to an unknown, underlying clean distribution $\mathcal{D}_{c}$.

In contrast, weakly supervised training utilizes a weakly labeled dataset $\mathcal{D}_{w}$ sampled from a noisy data distribution $\mathcal{D}_{n}$, represented as $\mathcal{D}_{n} = \{(x_1, \widehat{y}_1), (x_2, \widehat{y}_2), \dots, (x_N, \widehat{y}_N)\}$, where each label $\widehat{y}_{i}$ is derived from a weak labeling source and may differ from the actual or ground truth label $y_{i}$; the primary goal of weakly supervised training techniques is to develop a model $\mathcal{F}_{\theta}$ that can generalize well on the test dataset $\mathcal{D}_{te}$, assumed to be from a clean distribution $\mathcal{D}_{c}$, even though the model is trained solely on $\mathcal{D}_{w} \sim \mathcal{D}_{n}$.

To achieve this, we began with an internal dataset consisting of $\sim30K$ hours of unlabeled audios, covering various Arabic dialects—including Modern Standard Arabic (MSA)—as well as a diverse set of speakers, genders, and age groups. This dataset was subsequently filtered and labeled for training using the pipeline described in the following section, ending up with $\sim 15K$ hours of weakly annotated dataset. In addition to that, publicly available datasets were used as development and test sets for parameter optimization. Unlike the training data, these datasets provide clean transcriptions and serve as reliable ground truth references.

\subsection{Generation of Weak Labels}

Our labeling system consists of four main components, forming a structured pipeline illustrated in Figure \ref{fig:pipeline}. The process begins with the \textbf{Speech Processing Pipeline}, which segments the input audio while detecting overlapping speakers and filtering out low-quality segments. Each segment is limited to a maximum length of 5 seconds to minimize data loss if a segment is discarded.  

Next, the \textbf{Hypothesis Generators} produce
potential transcriptions for each audio segment, which are then evaluated by the \textbf{Hypothesis Selection Pipeline}. This module assigns scores to the generated hypotheses, filters out low-score segments, and selects the transcription for each segment based on the level of agreement between the transcriptions. More formally, given a set of hypotheses $\mathcal{H}$, the selected hypothesis $h_s$ is selected to minimize an error function $\varepsilon$ as shown in 
Equation \ref{eq:agreement}. Such that, the error function $\varepsilon(h_o, h)$ is defined as the Levenshtein distance \ref{eq:levenshtein}, where $d_L(h_o, h)$ represents the Levenshtein distance, which measures the minimum number of edits required to transform $h_o$ into $h$. Finally, the grammatical correctness of the selected hypothesis is evaluated by calculating its perplexity (PPL) with a language model (LM). If the hypothesis meets the quality threshold, it is retained; otherwise, the segment is discarded.

Lastly, to ensure diverse, short and long audio lengths, the \textbf{Merging Pipeline} combines consecutive small segments into larger chunks, while constraining the maximum audio chunk to 15 seconds. These chunks will serve as training data for our model training process.

\begin{equation}
    h_s = \arg\min_{h \in \mathcal{H}} \sum_{\substack{h_o \in \mathcal{H} \\ h_o \neq h}} \varepsilon(h_o, h)
\label{eq:agreement}
\end{equation}

\begin{equation}
    \varepsilon(h_o, h) = d_L(h_o, h) 
\label{eq:levenshtein}
\end{equation}

The complete labeling process is formally defined in Algorithm \ref{alg:dataset_labeling}. Initially, a proprietary ASR, FastConformer\footnote{\url{https://huggingface.co/nvidia/stt_ar_fastconformer_hybrid_large_pcd_v1.0}}, and Whisper \cite{pmlr-v202-radford23a} were used for hypothesis generation. However, after the first iteration, Whisper was excluded due to its tendency to produce hallucinated transcriptions. The full labeling pipeline was executed over two iterations to refine and enhance the dataset quality.

\begin{algorithm}
\label{alg:dataset_labeling}
\caption{Weak Label Generation Process}
\begin{algorithmic}[1]
\State \textbf{Input:} Dataset of audio files
\State \textbf{Output:} Labeled audio segments

\For{each audio file in dataset}
    \State Segment the audio using Voice Activity Detection (VAD)
    \For{each segment in audio}
        \If{duration of segment > threshold or segment has overlapping speakers}
            \State \textbf{Drop segment}
        \EndIf
    \EndFor
    \For{each remaining segment}
        \State Transcribe the segment using all selected ASR models
        \State Calculate Average Pairwise WER (Word Error Rate) and Average Pairwise CER (Character Error Rate) for each transcription
        \If{$PWER_{AVG}$ > $PWER_{threshold}$ or $PCER_{AVG}$ > $PCER_{threshold}$}
            \State \textbf{Drop segment}
        \Else
            \State Select the transcription with the highest agreement across models
                \If{$PPL$(Selected transcription) > threshold}
                    \State \textbf{Drop segment}
                \Else
                    \State Add the labeled segment to the dataset
                \EndIf
        \EndIf
    \EndFor
    \State Train a new ASR model on the labeled dataset
    \State Add the newly trained ASR model to the ASR list
\EndFor
\State \textbf{Repeat the pipeline on new data}
\end{algorithmic}
\end{algorithm}

Moreover, given the large number of hyperparameters involved in configuring the  pipeline, we formulate the selection process as an optimization problem. Our goal is to find the optimal set of hyperparameters that maximizes the amount of annotated data while preserving high annotation quality.

Let $\mathcal{A}$ denote the annotation pipeline, and let $\hbar$ be the set of hyperparameters used to configure it. When applied to an input audio $x$, the pipeline produces a set of annotated segments and their corresponding hypotheses:

\begin{equation}
    \mathcal{A}(x;\hbar) = \{(x_{s_i}, h_{s_i})\}_{i=1}^n
    \label{eq:pipeline}
\end{equation}

Here, $x_{s_i}$ denotes the $i^{th}$ sub-segment of $x$, and $h_{s_i}$ is its corresponding hypothesis. We define the sets of audio segments and hypotheses as:

\begin{equation}
    h_s = \{h_{s_i}\}_{i=1}^n,\quad x_s = \{x_{s_i}\}_{i=1}^n
    \label{eq:h_x}
\end{equation}

To evaluate the annotation efficiency, we define the efficiency score $\xi$ as the ratio between the total duration (or number of samples) of the annotated segments and the original input:

\begin{equation}
    \xi(x, x_s) = \frac{ \sum_{i=1}^{n} |x_{s_i}|}{|x|}
    \label{eq:efficiency}
\end{equation}

We define an error function $\mathcal{L}$ that quantifies the average discrepancy between the reference hypotheses $h_r$ and the predicted/selected hypotheses $h_s$ across all segments. The error is measured directly using the Word Error Rate (WER):

\begin{equation}
    \mathcal{L}(h_r, h_s) = \frac{1}{n} \sum_{i=1}^{n}\mathcal{E}(h_{r_i}, h_{s_i})
    \label{eq:overall-error}
\end{equation}

The optimization objective is to minimize the total error while maximizing annotation efficiency. We achieve this by jointly optimizing the hyperparameters $\hbar$ using a calibration set $\mathcal{D}_{\text{cal}}$, which, in this case, is the development set of the MASC dataset \cite{e1qb-jv46-21}.

\begin{equation}
 \max_{\hbar} \sum_{ \substack{(x,h_r) \in \mathcal{D}_{\text{cal}} \\ (x_s, h_s) \in \mathcal{A}(x; \hbar)}} \left[ \xi(x, x_s) - \mathcal{L}(h_r, h_s) \right]
 \label{eq:objective}
\end{equation}

This optimization framework ensures that the chosen hyperparameters yield a balance between high annotation accuracy and maximal coverage, leading to an effective and scalable annotation process.

\subsection{Model Architecture}

The Conformer model architecture \cite{gulati2020conformer} has become the de facto standard for automatic speech recognition due to its ability to capture both long-term and short-term dependencies in speech signals by employing multi-head self-attention and convolutional modules. Given its effectiveness, we adopt the Conformer architecture identical to that proposed in the original work. More precisely, we utilize the large variant of the model.

\section{Experiments and Results}
\label{sec:experiments}

\subsection{Experimental Setup}
In our experiment, we trained the Conformer model using Connectionist Temporal Classification (CTC) loss. To mitigate the limitations imposed by the conditional independence assumption inherent in CTC, we utilized a SentencePiece tokenizer. This tokenizer was specifically trained on our training data, with a vocabulary of 1024 unique tokens. The training process for the ASR model leveraged distributed data parallelism across 8 A100 GPUs, employing a global batch size of 512. The model's input features were 80-channel mel-spectrograms, which were computed using a 25-millisecond window and a 10-millisecond overlap.

For training, we employed the AdamW optimizer with $\beta_{1} = 0.85$ and $\beta_{2} = 0.97$, along with the Noam learning rate scheduler, using 10,000 warmup steps and a peak learning rate of $2 \times 10^{-3}$. For regularization purposes, we applied a dropout rate of 0.1 for all dropout modules and L2 regularization with a weight of $10^{-5}$. To speed up training and reduce memory consumption, we used bfloat16 precision. The model, which starts with weights initialized at random, comprises 18 Conformer layers. These layers feature a 512-dimensional space, are configured with 8 attention heads, employ a convolution kernel of size 31, and have a feedforward expansion factor of 4, resulting in approximately 121 million parameters.

\subsection{Evaluation Metrics \& Datasets}

We assessed the model using standard benchmarks, specifically Word Error Rate (WER) and Character Error Rate (CER). To ensure reproducibility and consistency, we adhered to the exact settings and codebase used in the Arabic ASR leaderboard \cite{wang2024open}\footnote{Code utilized from the main branch (commit c8c3a18095a071b7a86f73fb4ca7e66fa5e71180) of \url{https://github.com/Natural-Language-Processing-Elm/open_universal_arabic_asr_leaderboard}}. Consequently, we evaluated the model on the test sets of the same datasets, including SADA \cite{10446243}, Common Voice 18.0 \cite{commonvoice:2020}, MASC (clean and noisy) \cite{e1qb-jv46-21}, MGB-2 \cite{ali2016mgb}, and Casablanca \cite{talafha2024casablanca}. Table \ref{tab:datasets} summarizes the dialectal coverage of each dataset.

\begin{table}[htbp]
    \centering
    \label{tab:datasets}
    \begin{tabular}{l c l}
        \hline
        \textbf{Dataset} & \textbf{Dialects Count} & \textbf{Notes} \\
        \hline
        SADA \cite{10446243} & 10+ & Mostly Saudi dialect, includes MSA and 9+ other dialects \\
        Common Voice 18.0 \cite{commonvoice:2020} & 25 & Diverse dialect coverage \\
        MASC Clean \cite{e1qb-jv46-21} & 7 & Clean test set, includes MSA among other dialects \\
        MASC Noisy \cite{e1qb-jv46-21} & 14 & Noisy test set, includes MSA among other dialects \\
        MGB-2 \cite{ali2016mgb} & 4+ & ~70\% MSA, remaining are dialectal, also it includes English and French speech \\
        Casablanca \cite{talafha2024casablanca} & 8 & Includes multiple dialects with english, and french \\
        \hline
    \end{tabular}
    \vspace{1mm}
    \caption{Arabic leaderboard evaluation datasets and their dialect coverage.}
\end{table}

\subsection{Results}

The performance of our model was evaluated in terms of Word Error Rate (WER) and Character Error Rate (CER). Table \ref{tab:wer-results} presents the WER results, while Table \ref{tab:cer-results} details the CER achieved by our model and the baseline open source models. Employing greedy decoding without the integration of a language model, our model demonstrably outperforms all previously reported works. As shown in the tables, our proposed model consistently outperforms existing approaches under both clean and noisy conditions. It achieves the lowest average WER of 26.68 and an average CER of 10.05, compared to the best baseline model which records an average WER of 34.74 and an average CER of 13.37. Beyond that, the results detailed in Table \ref{tab:closed-source-models-result} compare the performance of our model against a closed-source alternative, clearly indicating the superior outcomes obtained with our proposed method. These results highlight the robustness of our model.

Furthermore, Table \ref{tab:reduction-results} quantifies the relative improvement of our model over the strongest baseline in terms of Word Error Rate Reduction (WERR) and Character Error Rate Reduction (CERR). Our model achieves a significant average reduction of 23.19\% in WER and 24.78\% in CER.

\begin{table}[H]
\centering
\resizebox{\textwidth}{!}{
\begin{tabular}{|c|c|c|c|c|c|c|c|}
\hline
 & SADA & Common Voice & MASC (clean) & MASC (noisy) & Casablanca & MGB-2 & \textbf{Average} \\ \hline
Facebook Seamless m4t Large V2 & 62.52 & 21.7 & 25.04 & 33.24 & 66.25 & 20.23 & 38.16 \\ \hline
OpenAI Whisper Large V3 & 55.96 & 17.83 & 24.66 & 34.63 & 71.81 & 16.26 & 36.86 \\ \hline
OpenAI Whisper Large V2 & 57.46 & 21.77 & 27.25 & 38.55 & 71.01 & 25.17 & 40.20 \\ \hline
OpenAI Whisper Large & 63.24 & 26.04 & 28.89 & 40.79 & 72.18 & 24.28 & 42.57 \\ \hline
OpenAI Whisper Large V3 Turbo & 60.36 & 25.73 & 25.51 & 37.16 & 73.79 & 17.75 & 40.05 \\ \hline
OpenAI Whisper Medium & 67.71 & 28.07 & 29.99 & 42.91 & 75.44 & 29.32 & 45.57 \\ \hline
OpenAI Whisper Small & 78.02 & 24.18 & 35.93 & 56.36 & 87.64 & 48.64 & 55.13 \\ \hline
Nvidia Parakeet CTC 1.1B Concat & 70.70 & 26.34 & 30.49 & 45.95 & 80.80 & 24.94 & 46.54 \\ \hline
Nvidia Parakeet CTC 1.1B Universal & 73.58 & 40.01 & 36.16 & 50.03 & 81.30 & 30.68 & 51.96 \\ \hline
Nvidia Conformer CTC Large V3 & 47.26 & 10.60 & 24.12 & 35.64 & 71.13 & 19.69 & 34.74 \\ \hline
\textbf{Ours} & \textbf{27.71} & \textbf{10.42} & \textbf{21.74} & \textbf{28.08} & \textbf{60.04} & \textbf{12.10} & \textbf{26.68} \\ \hline
\end{tabular}
}
\vspace{0.01pt}
\caption{Word Error Rate (WER) comparison of different open-source models, demonstrating the superior performance of our model across all datasets.}
\label{tab:wer-results}
\end{table}

\begin{table}[H]
\centering
\resizebox{\textwidth}{!}{
\begin{tabular}{|c|c|c|c|c|c|c|c|}
\hline
 & SADA & Common Voice & MASC (clean) & MASC (noisy) & Casablanca & MGB-2 & \textbf{Average} \\ \hline
Facebook Seamless m4t Large V2 & 37.61 & 6.24 & 7.19 & 11.92 & 29.85 & 9.37 & 17.03 \\ \hline
OpenAI Whisper Large V3 & 34.62 & 5.74 & 7.24 & 12.89 & 35.04 & 7.74 & 17.21 \\ \hline
OpenAI Whisper Large V2 & 36.59 & 7.44 & 8.28 & 15.49 & 36.00 & 13.48 & 19.55 \\ \hline
OpenAI Whisper Large & 40.16 & 9.61 & 9.05 & 16.31 & 35.71 & 12.10 & 20.49 \\ \hline
OpenAI Whisper Large V3 Turbo & 37.67 & 10.89 & 7.55 & 13.93 & 34.83 & 8.34 & 18.87 \\ \hline
OpenAI Whisper Medium & 43.83 & 10.38 & 8.98 & 17.49 & 38.12 & 14.82 & 22.27 \\ \hline
OpenAI Whisper Small & 33.17 & 6.79 & 9.01 & 19.43 & 46.12 & 15.56 & 21.68 \\ \hline
Nvidia Parakeet CTC 1.1B Concat & 46.70 & 9.82 & 8.41 & 18.72 & 49.77 & 9.87 & 23.88 \\ \hline
Nvidia Parakeet CTC 1.1B Universal & 49.48 & 14.64 & 10.29 & 20.09 & 45.31 & 11.36 & 25.19 \\ \hline
Nvidia Conformer CTC Large V3 & 22.54 & \textbf{3.05} & \textbf{5.63} & 11.02 & 30.5 & 7.46 & 13.37 \\ \hline
\textbf{Ours} & \textbf{11.65} & 3.21 & 5.80 & \textbf{8.88} & \textbf{25.51} & \textbf{5.27} & \textbf{10.05} \\ \hline
\end{tabular}
}
\vspace{0.01pt}
\caption{Character Error Rate (CER) comparison across different open-source models. Our model demonstrates strong performance on almost all datasets and achieves comparable results on Common Voice and MASC (noisy).}
\label{tab:cer-results}
\end{table}

\begin{table}[htbp]
\centering
\resizebox{\textwidth}{!}{ 
\begin{tabular}{|c|cc|cc|cc|cc|cc|cc|cc|}
\hline
\multirow{2}{*}{} & \multicolumn{2}{c|}{SADA} & \multicolumn{2}{c|}{Common Voice} & \multicolumn{2}{c|}{MASC (clean)} & \multicolumn{2}{c|}{MASC (noisy)} & \multicolumn{2}{c|}{Casablanca} & \multicolumn{2}{c|}{MGB-2} & \multicolumn{2}{c|}{\textbf{Average}} \\ \cline{2-15} 
                  & \multicolumn{1}{c|}{WER} & CER & \multicolumn{1}{c|}{WER} & CER & \multicolumn{1}{c|}{WER} & CER & \multicolumn{1}{c|}{WER} & CER & \multicolumn{1}{c|}{WER} & CER & \multicolumn{1}{c|}{WER} & CER & \multicolumn{1}{c|}{WER} & CER \\ \hline

{ElevenLabs Scribe}        & \multicolumn{1}{c|}{49.44} & 23.33 & \multicolumn{1}{c|}{{28.27}} & 7.33 & \multicolumn{1}{c|}{{31.93}} & 8.23 & \multicolumn{1}{c|}{{41.23}} & {13.14} & \multicolumn{1}{c|}{{63.77}} & {27.17}  & \multicolumn{1}{c|}{{25.68}} & {9.27} & \multicolumn{1}{c|}{{40.05}} & {14.74} \\ \hline

{Microsoft Azure STT\footnotemark}        & \multicolumn{1}{c|}{58.5} & 35.39 & \multicolumn{1}{c|}{33.77} & 9.29 & \multicolumn{1}{c|}{40.66} & 14.73 & \multicolumn{1}{c|}{45.64} & 15.77 & \multicolumn{1}{c|}{64.84} & 27.84  & \multicolumn{1}{c|}{30.91} & 13.7 & \multicolumn{1}{c|}{45.72} & {19.45} \\ \hline

{OpenAI GPT-4o transcribe}        & \multicolumn{1}{c|}{66.47} & 49.57 & \multicolumn{1}{c|}{28.19} & 8.14 & \multicolumn{1}{c|}{31.53} & 8.85 & \multicolumn{1}{c|}{43.29} & 18.81 & \multicolumn{1}{c|}{70.72} & 43.15  & \multicolumn{1}{c|}{29.62} & 17.34 & \multicolumn{1}{c|}{44.97} & {24.31} \\ \hline

\textbf{Ours}        & \multicolumn{1}{c|}{\textbf{27.71}} & \textbf{11.65} & \multicolumn{1}{c|}{\textbf{10.42}} & \textbf{3.21} & \multicolumn{1}{c|}{\textbf{21.74}} & \textbf{5.80} & \multicolumn{1}{c|}{\textbf{28.08}} & \textbf{8.88} & \multicolumn{1}{c|}{\textbf{60.04}} & \textbf{25.51}  & \multicolumn{1}{c|}{\textbf{12.10}} & \textbf{5.27} & \multicolumn{1}{c|}{\textbf{26.68}} & \textbf{10.05} \\ \hline

\end{tabular}
}
\vspace{0.01pt}
\caption{WER and CER comparison of closed-source models, highlighting the superior performance of our model across all benchmark datasets.}
\label{tab:closed-source-models-result}
\end{table}

\footnotetext{Given the numerous Arabic Automatic Speech Recognition (ASR) models available on Azure for each dialect, evaluating every model was impractical. Therefore, we selected two representative models, the SA and JO models, and averaged their performance metrics. Furthermore, these two models exhibited nearly identical WER and CER in our initial comparison.}

\begin{table}[H]
\centering
\resizebox{\textwidth}{!}{ 
\begin{tabular}{|cc|cc|cc|cc|cc|cc|cc|}
\hline
\multicolumn{2}{|c|}{SADA}      & \multicolumn{2}{c|}{Common Voice} & \multicolumn{2}{c|}{MASC (clean)} & \multicolumn{2}{c|}{MASC (noisy)} & \multicolumn{2}{c|}{Casablanca} & \multicolumn{2}{c|}{MGB-2}     & \multicolumn{2}{c|}{Average}   \\ \hline
\multicolumn{1}{|c|}{WERR} & CERR & \multicolumn{1}{c|}{WERR}   & CERR  & \multicolumn{1}{c|}{WERR}   & CERR  & \multicolumn{1}{c|}{WERR}   & CERR  & \multicolumn{1}{c|}{WERR}  & CERR & \multicolumn{1}{c|}{WERR} & CERR & \multicolumn{1}{c|}{WERR} & CERR \\ \hline

\multicolumn{1}{|c|}{\textbf{41.36}}    &  \textbf{48.31}   & \multicolumn{1}{c|}{\textbf{1.69}}      &   -5.24   & \multicolumn{1}{c|}{\textbf{9.86}}      &   -3.01   & \multicolumn{1}{c|}{\textbf{15.52}}      &  \textbf{19.41}    & \multicolumn{1}{c|}{\textbf{5.84}}     &   \textbf{6.10}  & \multicolumn{1}{c|}{\textbf{25.58}}    & \textbf{29.35}    & \multicolumn{1}{c|}{\textbf{23.19}}    &  \textbf{24.78}   \\ \hline
\end{tabular}
}
\vspace{0.01pt}
\caption{Comparison of Word Error Rate Reduction (WERR) and Character Error Rate Reduction (CERR) achieved by our model relative to the best performing baseline. Our model demonstrates an average reduction of 23.19\% in WER and 24.78\% in CER when compared with the best performing baseline model.}
\label{tab:reduction-results}

\end{table}

\section{Conclusion}
\label{sec:conclusion}

In this work, we presented a novel approach to Arabic Automatic Speech Recognition (ASR) by leveraging weakly supervised learning to address the challenges posed by limited labeled datasets. Our method achieves state-of-the-art (SOTA) performance across both Modern Standard Arabic and dialectal speech, establishing a new benchmark for Arabic ASR. By demonstrating the effectiveness of weakly supervised learning, we provide a scalable and cost-effective alternative to traditional supervised approaches, significantly reducing reliance on large annotated datasets. This advancement not only contributes to the development of Arabic ASR but also offers a promising direction for enhancing ASR systems in other low-resource languages.

\bibliographystyle{unsrt}  
\bibliography{references}

\end{document}